# Establishment of Relationships between Material Design and Product Design Domains by Hybrid FEM-ANN Technique

K. Soorya Prakash [1], S. S. Mohamed Nazirudeen [2] and M. Joseph Malvin Raj [3]

[1,3] Department of Mechanical Engineering,
Anna University Coimbatore, Tamil Nadu, India

[2] Department of Metallurgical Engineering,
PSG College of Technology, Coimbatore, Tamil Nadu, India.

**Abstract**
In this paper, research on AI based modeling technique to optimize development of new alloys with necessitated improvements in properties and chemical mixture over existing alloys as per functional requirements of product is done. The current research work novels AI in lieu of predictions to establish association between material and product customary. Advanced computational simulation techniques like CFD, FEA interrogations are made viable to authenticate product dynamics in context to experimental investigations. Accordingly, the current research is focused towards binding relationships between material design and product design domains. The input to feed forward back propagation prediction network model constitutes of material design features. Parameters relevant to product design strategies are furnished as target outputs. The outcomes of ANN shows good sign of correlation between material and product design domains. The study enriches a new path to illustrate material factors at the time of new product development.

## 1. Introduction

Products designed for recognized applications with problem-solving design, expected quality, believable life is of high demand and has become a mortal of human living. Purchasing a material with a restricted chemical composition and defined mechanical properties and utilizing at conceptual stage of product development is the present scenario. For the selected material, the required shape (casting, forging, plate, and pipe) and the possible manufacturing processes should be considered first. Then in accordance with the design code (e.g., ASTM, ANSI, BIS) adopted, the specification for the material to be applied is selected and the allowable design stresses at the operation temperature, are determined. In correlation with design procedure, the proved uneconomical process in terms of cost and time namely physical model or prototype forms the basis for better understanding of a particular process. Furthermore, it is very difficult to develop prototypes for complex and multi dimensional products and then subjecting the same for property and functionality testing [5]. In times of yore, development of new material right from concept to the entire type of application including low intensity services and sky-scraping services like aero space applications, missiles, gas turbines, high pressure areas etc., has taken typically more than ten years for complete development.

A designed component is a unique part or subassembly that must be specifically designed and fabricated as part of the design being developed. Choosing between designed and standard components can be an important consideration. The choice between designing a special purpose component, optimized for performance and weight, or purchasing a standard, "off-the-shelf" component from a supplier, can have far reaching performance, cost, quality and timing consequences. Market availability is lofty for multiple design build test cycles involved in the creation of a design. The product life depends highly upon the type and property of the material and in turn the properties are forbidden for the effective functionality.

It is obvious that chemistry of the alloy plays a vital role in determination of material parameters. As of functionality of the product is concerned, material characterization attributes involves the complex interrelation of a number of variables associated with product design, its manufacturing process and service conditions. In cases where design is typically too risky to develop , a first round of planning, material selection, functional design, manufacture and assembly is undertaken to produce a model, which when tested may indicate ways of improving the product design.

The current study is focused on imparting this information in conjunction with possibly extensive computer aided analysis and optimization either to manufacture an improved model for tests or to manufacture a prototype or the product itself. On thorough review over various literatures it is proved to be worthwhile to blend relationships between the domains of material design and product design inspite of various other considerable governing factors of design.

## 3. Experimental Schema

Experiments are conducted in a foundry for the ASTM A487 Gr 4C: 0.2 to 0.3 % Carbon; 0.4 to 0.8 % Silicon; 0.8 to 1.0 % Manganese; 0 to 0.03 % Sulphur; 0 to 0.03 % Phosphorus; 0.4 to 0.8 % Chromium; 0.4 to 0.8 % Nickel; 0.15 to 0.3 % Molybdenum; maximum 0.5% Copper; 0 to 0.03 % Vanadium; 0 to 0.1 % Tungsten and a maximum of 0.60% of unspecified alloying elements. The melting range of the specified steel is about 2740-28000F. The test castings were poured in $CO_2$ – sodium silicide moulds, knocked out, risers were removed and then subjected to normalising, tempering sort of heat treatments. A sample of the data set is made available in table 1.

Table1: Set of sample data used for model development

| Sl No | Chemical Composition ( in % ) | | | | | | | | | | | Mechanical Properties | | | | | Test temp °C |
|---|---|---|---|---|---|---|---|---|---|---|---|---|---|---|---|---|---|
| | C | Si | Mn | S | P | Cr | Ni | Mo | Cu | V | W | TS | YS | El | RA | IS | |
| 1 | 0.26 | 0.51 | 0.81 | 0.011 | 0.014 | 0.54 | 0.54 | 0.24 | 0.03 | 0.002 | 0.005 | 639 | 489 | 23.0 | 45.8 | 42 | -46 |
| 2 | 0.23 | 0.55 | 0.84 | 0.010 | 0.015 | 0.50 | 0.45 | 0.18 | 0.03 | 0.003 | 0.008 | 635 | 509 | 22.6 | 36.0 | 58 | -46 |
| 3 | 0.24 | 0.63 | 0.91 | 0.021 | 0.022 | 0.51 | 0.46 | 0.19 | 0.04 | 0.003 | 0.010 | 646 | 451 | 20.0 | 45.0 | 34 | -46 |

## 2. Scope of Research

For this purpose artificial neural network has evolved as an efficient tool for developing models which costs less compared to traditional process [15]. To generate a new material with multiplicity in composition is too tedious and this proves to be a barrier for the development of many new products. As a remedy for this deficient and headed for the formulation of innovative materials, survival of ground-breaking procedures is a must. The properties used are those involved in the basic formulas of mechanical design and those typically used to characterize material properties provided in engineering handbooks, specifications and supplier catalogs. Designers are typically interested in strength, ductility, toughness, Young's modulus and fatigue strength. These models states that the combination of certain properties may impose performance limitations in addition to those constraints imposed by individual properties.

The scope of this research on artificial intelligence based modeling technique is to optimize the ASTM A487 Gr 4C high pressure application material with necessitated improvements in properties and chemical mixture of the existing alloys as per the functional requirements of the valves as shown in figure 1.

Behavior of material depends on percentage composition of alloying elements and so the same is projected out for 250 tests wherin odd experimental observations with missing or contradictory interpretations and of abnormal strength values were omitted from the training data set.

The material attributes like Yield Strength (YS) in Mpa Tensile Strength (TS) in Mpa, Percentage Elongation (El), Percentage Reduction in Area (RA), and Impact Strength (IS) of the specimens were experimented. In routine, utmost care is taken to maintain the test temp at -46°C (for impact testing). Equipped with the CHARPY Impact testing machine for test specimens prepared as per the standard (ISO CVN 2 mm SIZE - 10 x 10 x 55 mm) the impact tests are accomplished.

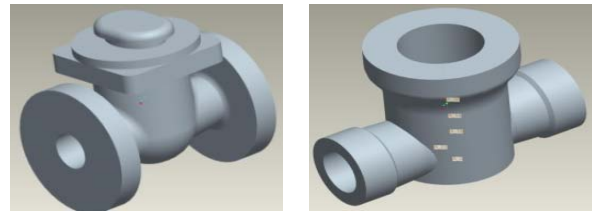

Fig. 1(a) & 1(b) shows the solid model geometry for Gate & Ball Valve respectively

3.1 Computational Fluid Dynamic Analysis

In order to understand and identify the thrust areas of high pressure distribution, the modeled components are imported to Fluent, software known for Computational Fluid Dynamics analysis [9]. In analysis the pressure distribution across the required area is analyzed by gratifying the boundary condition applied. Table 2 clearly pictures that two types of valves viz Ball valve and gate valve are subjected to 3 type of fluids.

Table 2: Inlet boundary conditions for CFD

| Fluid Parameters | | Water | Lubricant | Diesel |
|---|---|---|---|---|
| Density (kg/m$^3$) | | 951 | 875 | 834 |
| Viscosity (cp) | | 1 | 22.2 | 4 |
| Inlet pressure (bar) | | 350 | 280 | 180 |
| Temperature ($^0$C) | | 110 | 100 | 15.6 |
| Dia$_{in}$ (mm) | Valve$_1$ | 87 | 87 | 87 |
| | Valve$_2$ | 51 | 51 | 51 |
| Dia$_{out}$ (mm) | Valve$_1$ | 54 | 54 | 54 |
| | Valve$_2$ | 75 | 75 | 75 |

The CFD code of fluent 6.0 for finite volume method has been utilized to solve the discretization of continuity equation and Navier stokes equation. The CFD code is commonly used to solve since it has high capability of solving the transient, compressible, turbulent and reacting flows in the finite volume grid with boundary condition and meshes [2].

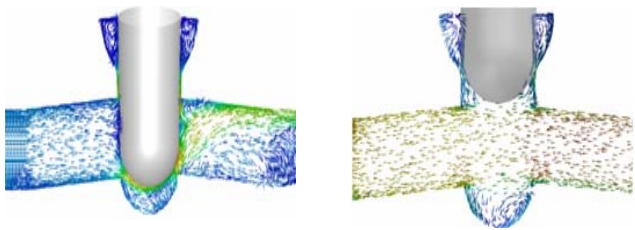

Fig. 2(a) & 2(b) shows sample velocity vector figures of Gate valve for 10mm and 70 mm opens respectively

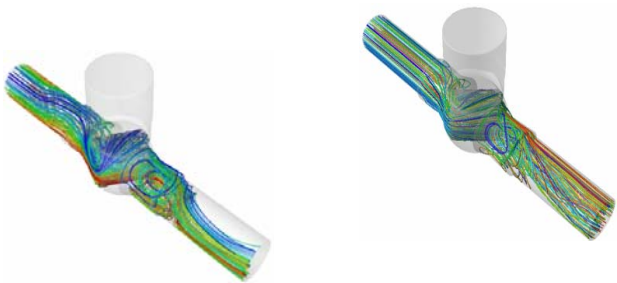

Fig. 3(a) & 3(b) shows sample fluid flow regions of Ball valve for 30 degree and 45 degree opens respectively

On thorough analysis it is found that the maximum pressure is attained in the fluid flow viz. Water for both valves. The same is attained since the viscosity of water is on the lower side when compared to any other fluid considered in this study. The valve body manufactured by this particular ASTM Grade A487 material has the ability to withstand a maximum dynamic pressure of 330 bar approximately. The results obtained through computational fluid dynamics performed over ball and gate valve is shown clearly in figures 1 & 2.

3.2 Finite Element Analysis

As future prospects, the ability of the proposed valve material in relation to fatigue characteristics [12] is identified using Finite Element Analysis by applying the observed results as input parameters in adoption to variation in thickness of ball valve and outer diameter of gate valve. The same is carried over to assess the stress concentration and to verify the contact status between the ball and the valve body finite element method is used. Assessments of the stress concentration, strain distribution, total deformation acting on the ball and gate valves for its different valve position is carried out. The output results for both valve types credited through finite element analysis are illustrated in figures 4 & 5.

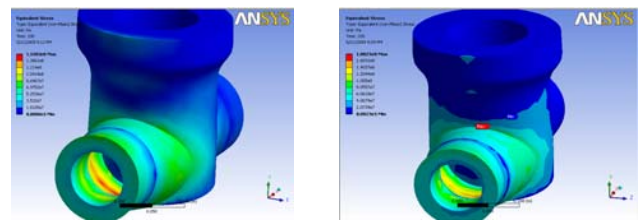

Fig 4 (a) & 4(b) shows sample analysis results for Ball Valve of 21 mm thickness 30 & 45 degree opening modes respectively

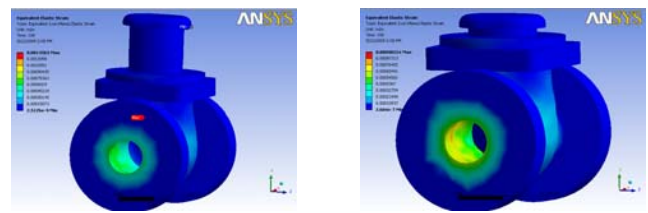

Fig 5 (a) & 5 (b) shows sample analysis results for Gate valve of 90 mm OD for 10 mm and 70 mm openings respectively

It is found that the maximum stress is achieved over minimum thickness valve geometry. The maximum stress in the regions around the inner walls is approximately 4.3302*108 Pascal; maximum strain is 0.002165 m/m, minimum life of 2116.1 cycles and total deformation of 5.92 *10-5 was found to be allowable in this present study. In view of the fact that this research [13,14] is highly oriented towards the prediction of life for the optimum geometry of a selected product, graphs for life and poisons ratio is figured out and is as shown in fig 6.

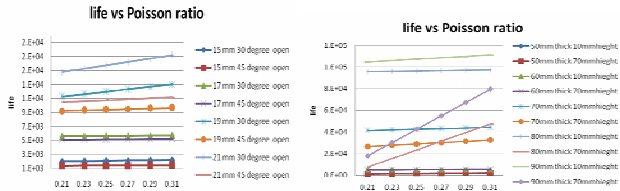

Fig. 6 shows the graph of Life vs. Poisson ratio for (a) Ball Valve of 30 & 45 degree opening  (b) Gate valve of 10 & 70 mm opening heights

Henceforth on effective supplementation of specified input parameters, the allowable thickness, the acceptable compositions and permissible life of the casting design can be achieved with integration of experiential results with Artificial Intelligence.

## 4. Modeling Technique

Neural networks are parameterized non-linear models used for empirical regression and classification modeling. Their inherent flexibility enables them to discover more complex relationships in data than traditional models. The structure of the network, as shown in Fig. 1, consists of 16 numbers of inputs (experimental variables), 5 outputs and an intermediate hidden layer. The specification of the network structure together with the set of weights, gives a complete description of the formula relating the inputs to the output [8]. The weights are a set of coefficients determined by 'training' the neural network on a set of input and output data. The training process involves the minimization of a regularized sum of square errors, which leads to the optimum non-linear relationship between inputs and outputs. This process is quite computer intensive. But, once the network is trained, the estimation of outputs for any given set of inputs is very rapid. For the proposed study the network as shown in figure 8, all 11 elements of composition and their relative 4 mechanical properties along with the thickness of the valve body are considered to be the inputs. Strictly speaking the necessitated parameters of material design are considered to be the inputs for the network developed through artificial intelligence procedure. As specified the scope of the proposed study was to blend a relationship between material design and product design domains. Hence the outputs specified were highly based on the product design parameters [17]. In total, 5 outputs namely stress, strain, total deformation, life and years of service are considered. The number of hidden units decides the complexity of a model. Bearing this concept of ANN, various no of hidden layers and respective number of neurons are tried over. On careful observation over theories and on practical test conducted for the given set of data's, optimal number of hidden layers subjected for training this ANN network was found to be 2. On thorough review over cluster of literatures in relation to ANN based material design, it was concluded that among the various types of networks available, feed forward back propagation type of network was supposed to be the best suitable for this application. A feed-forward back propagation network as shown in Figure 8 has been used here as a simulator to find an optimum mapping between inputs and outputs. The theory behind back propagation rule has been well documented in many publications. Despite the fact that the same dependent and independent variables are used in this research, different ANN models were developed and have been estimated for carbon and low alloy standard specified steel castings suitable for pressure service [3].

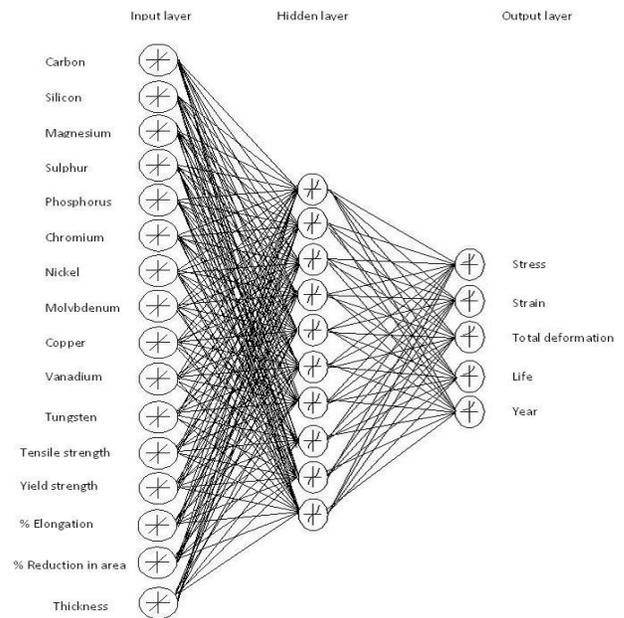

Fig. 8  Illustrates typical network structure used in the analysis.

On having decided the network type, the network was trained over combination of various other paradigms of artificial intelligence network like training function, adaption learning function and performance function on trial and error basis. After periodical observations over these paradigms when subjected to training, the following was selected to be optimum for this particular application of 16 material design inputs and 5 product design outputs:

| Training function | traingdm |
| Adaption learning function | learngdm |
| Performance function | MSREG |

Above all, maximum possible permutation and combinations of the above said paradigms was tried out and the same was generalized to be the best functions. The number of hidden layers was projected to be 2. Wherein for the finalized network the number of hidden neurons was predetermined to be 10 for layer 1. To have a better report on the overall network performance, the ANN network was trained for tansig & log sig modes of transfer functions as far as layer 1 is concerned. Whereas the number of hidden neurons for layer 2 was specified as 5 and permanently transfer function purelin was imposed for both type of analysis. This adhesion step of different transfer functions for hidden layers was again constituted by conducting real time training of network for all possible considerations.

However, an overly complex model generalizes poorly and fits even the noise in the data rather than the underlying trend. This phenomenon is known as 'over fitting' and can be avoided by dividing the database into a training set and a test set. The network is first trained on the training set and then its performance is tested using the test set to confirm the robustness of trained ANN. Networks that produce curves adequately fitting both the training and the test set are generally chosen for making predictions on unseen data.

Records associated with materials, which exhibit unusual property values were removed from the dataset. When network training was restarted, the performance of the network improved considerably, allowing accurate generalized predictions of the relative domains. The experimental data were segregated into three different categories. First set of 95 experimental data was used for training the network model so as to adapt to the present work on new material design i.e. to predict the property for a given set of new composition values. The second set of 40 experimental data's was used for testing the network model developed. And the third set of 11 experimental data's was utilized for validation purpose.

The trained neural networks have been used to predict the properties of the materials in the test datasets which have then been compared to the experimental results.

The network topology (i.e., the number of the input variables, the number and size of the hidden layers), was a critical parameter affecting the performance of the model. The topology of the developed ANNs were determined with repetitive trials, keeping the architectures as simple as possible to avoid overtraining due to the limited number of samples available in both training and test sets. The model with the smallest prediction error on the test data set after training was used to predict the test set and to verify the accuracy of the model for all ranges of thicknesses.

## 5. Results and discussion

The software MATLAB is used to develop the ANN model to suit the research requirement of material design and product design integration, algorithms were developed. In property prediction researches, the training algorithm is used to iteratively adjust the network's interconnection weights so that the error in prediction of the training dataset records is minimized and the network reaches a specified level of accuracy. The network can then be used to predict output values for new input data and is said to generalize well if such predictions are found to be accurate. On a trial and error basis various epoch levels and goals has been set initially for predicting the specified mechanical property. Also for effective validation of the results, the input-output patterns are normalized before training. By the use of these normalized data's, the network is been triggered for different mechanical properties specified under various combinations of training functions in order to attain a stabilized network [8].

6.1 Prediction Performance of the Network Trained for 15 mm Dataset

The prediction performance of the developed network was subjected to training for 15 mm thickness data sets. The performance goal for the network was set to zero. As prescribed earlier, the network was trained for tansig and logsig type of transfer functions. The same was carried out as better results were yielded during the initial validation of network. The figure numbered 9 and 10 clearly illustrates the initial performance of the model. Similarly the figure 11 and 12 exemplifies the final outcomes of both transfer functions. On thorough consideration of both these cases, it was identified that in case of tansig initial peaks was observed. But the final results for both the transfer functions did not yield much difference. Also from final performance figures 11 & 12 it can be clearly witnessed that performance level becomes constant before reaching zero value. This strategy means that the network has been fully trained and further training is not possible. The Y-axis representation in Figures defines the performance value. Performance ratio was distorted with change in values and by iterations, 0.1 fits best to this particular application.

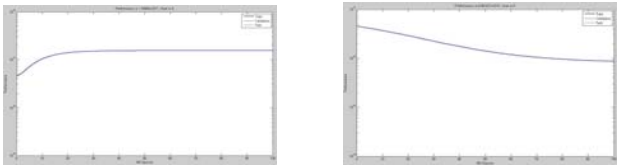

Fig 9 Initial Network performances at 100 epochs for 15 mm thickness using logsig & tansig as transfer functions respectively.

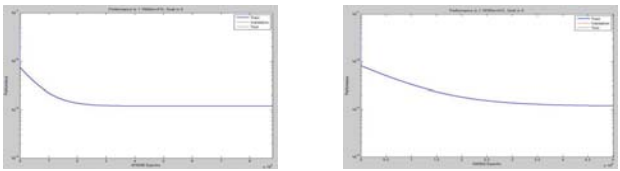

Fig. 10 Final Network performances at 879300 & 500050 epochs for 15 mm thickness using logsig& tansig as transfer functions respectively.

The number of epochs taken to reach saturation upon training was more in case of logsig transfer function when compared with tansig transfer function. The performance level was considerably more in case of logsig when performed for 15 mm thickness of the valve body.

6.1 Prediction Performance of the Network Trained for 17 mm Dataset

The developed ANN network was constituted to training for 17 mm thickness data sets. Prearranged datas for composition and property along with 17 mm constant thickness was given as input. The performance goal for the network was set to zero. As prescribed earlier, the network was trained for tansig and logsig type of transfer functions.. Figures 13 and 14 demonstrate the initial performance of the network model for logsig and tansig mode of transfer function. Likewise figure 15 typify the final Network performances at 1268951epochs for 17 mm thickness using logsig as transfer function and figure 16 represent the final outcomes at 1280176 epochs for 17 mm thickness using tansig as transfer function. As in previous case the network has been fully trained before reaching the goal and further training was stopped.

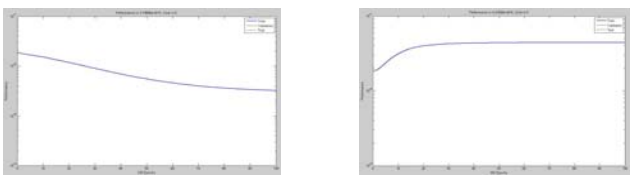

Fig. 11 Initial Network performances at 100 epochs for 17 mm thickness using logsig & tansig as transfer functions respectively.

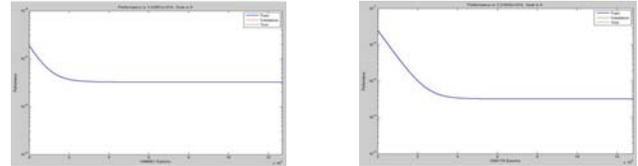

Fig. 12 Final Network performances at 1268951 & 1280176 epochs for 17 mm thickness using logsig & tansig as transfer functions respectively.

The number of epochs taken to reach saturation upon training was almost equal in both the cases of transfer function when performed for 17 mm thickness of the valve body.

6.1 Prediction Performance of the Network Trained for 19 mm Dataset

The network was reinitialized for training for 19 mm thickness data sets. The performance goal for the network was set to zero. As given earlier, the adopted network was trained for tansig and logsig transfer functions. The performance of the developed model for these data sets on initial training could not yield good results. The figures 17 and 18 picture the initial performance of the model at 100 epoch level. Upon further rigorous training of the network model, for being the goal set to zero, remarkable variations in training was observed. Unaltered performance could be observed at the early stage of the training. But training was continued until a maximum epoch of approximately 1100000 was reached. Very minimal down fall of curve, if not improvement in network performance as depicted in figure 19 and 20 was visualized at the time of stopping the network for the specified goal and epochs. The network performance rate was better in case of tansig transfer function in this case.

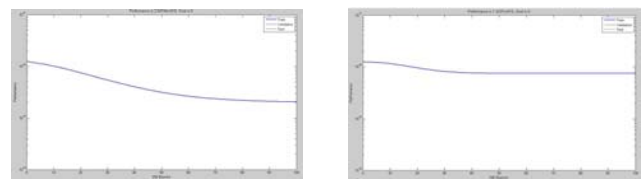

Fig. 13 Initial Network performances at 100 epochs for 19 mm thickness using logsig & tansig as transfer functions respectively.

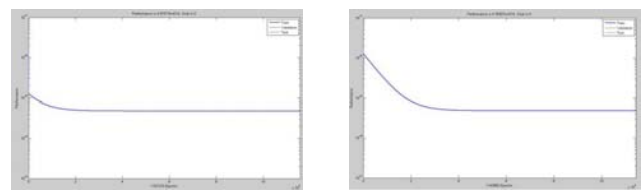

Fig. 14 Final Network performances at 1167275 & 1143960epochs for 19 mm thickness using logsig & tansig as transfer functions respectively.

## 6.1 Prediction Performance of the Network Trained for 21 mm Dataset

Initial training for 21 mm thickness data sets was made through for goal being set to zero and the epoch level 100. Clear examination on the training flow intimates that for logsig transfer function, steady declination in curve to reach the goal was targeted from initial state. Figure 21 and 22 point up the theory of network performance. But when the same network was trained for higher epoch levels and for 21 mm data sets, finer results were attained for tansig transfer functions. Deep and steady flow of graph can be envisaged through figure 24. For this particular data sets stabilized training outcomes was obtained at 700000 epochs approximately. Upon further training of the network, considerable change performance could not be seen and hence the network was ended at these early epoch levels for 21 mm thickness of the valve body..

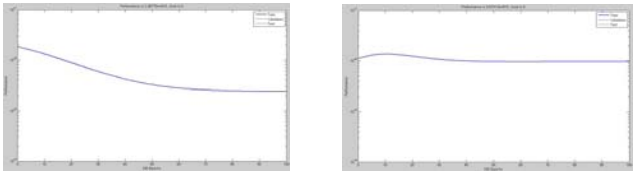

Fig. 15 Initial Network performances at 100 epochs for 21 mm thickness using logsig & tansig as transfer functions respectively.

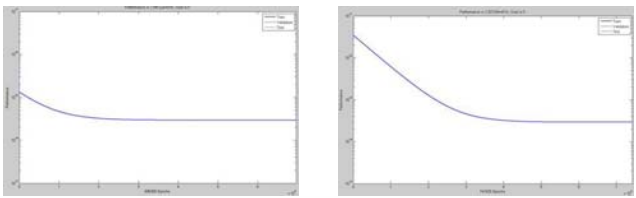

Fig. 16 Final Network performances at 696300 & 741925 epochs for 21 mm thickness using logsig & tansig as transfer functions respectively.

## 7. Testing and Validation

The theory of artificial intelligence utters an inevitable fact that the data used for testing and validation should not be included in training circumstances. As a first step the network has to be triggered for training the data sets by comparing experimental results and predicted outcomes. The next step in evaluation is, to test the Back-Propagation network performance with the other test samples that were never applied to network before [5]. Hence the network models developed by ANN were subjected to testing with 40 newer data sets collected for this purpose and their results are to be predicted. The weights obtained through training process were supplied as initial weights for validation purpose. Even though the desired goal set to zero was not reached, the network model developed yielded better results.

Upon the effective completion of validation, the initial weights were reverted with weights observed through training of the network model developed for testing purpose also. In order to check the robustness of the Back-Propagation network controller 11 input data sets were fed into the trained network. A detailed analysis on the tested data as per the theory of ANN [3,6,8] the obtained results are well within the predictable ranges.

After training the network successfully, it has been tested by using the known test data. Statistical methods were used to compare the results produced by the network. Errors occurring at the learning and testing stages are called the root-mean squared (RMS), absolute fraction of variance ($R^2$), and mean error percentage values. These are defined in many literatures [6,17,18] and are as follows, respectively:

$$\text{RMS} = \left( (1/p) \sum |t_j - o_j|^2 \right)^{1/2}, \quad (1)$$

$$R^2 = 1 - \left( \frac{\sum_j (t_j - o_j)^2}{\sum_j (o_j)^2} \right), \quad (2)$$

$$\text{mean \% error} = \frac{1}{p} \sum_j \left( \frac{t_j - o_j}{t_j} 100 \right). \quad (3)$$

The prediction errors were achieved on calculations using the equations 1, 2, 3 to account the performance of the ANN network developed for blending design domains. There is no definite rule as how the training and prediction errors correlate [17, 18]. The RMS error for the training data is reduced as the training continues though the improvement of some iteration becomes very small and sometimes RMS error may begin to increase.

## 8. Conclusion

In this study, finite-element modeling (FEM) and artificial neural networks (ANN) analysis were combined to establish relationship between material design and product design domains. As a result of this study, it was possible to determine the optimal parameter settings for training the given network. On the other hand, the network developed through artificial intelligence approach detects the relationships of the presented data during its training phase and updates itself in such way as to map best the properties of the phenomenon presented to it.

In this present work, different transfer functions have been investigated to get accurate results. Different combinations of governing parameters like performance ratio, number of epochs etc. were tried over a range of accepted values. The basic phenomenon of the prediction model developed is to give information on the properties for the specified material and to continue with the process to go ahead with next phases as usually. Hence the study ascertains a robust affiliation between the design variables of material design domain and product design domain.

Finally it is been ensured that the algorithm developed for has the ability to produce all necessary appropriate results well within the prescribed RMS error. The neural network approach appears to be a very powerful tool in materials engineering. Study on the error results enables that the prediction of the product variables in context to material variables of the considered steel is in a good agreement with the experimental data.

K.Soorya Prakash is currently working as faculty in the Department of Mechanical Engineering at Anna University, Coimbatore. He has completed his post graduation in Production engineering and pursing his doctoral degree. His research interest is mainly focused on identification and analysis of materials for newer products and emerging processes. He has to his credit many national and international Publications.

Dr. S.S. Mohamed Nazirudeen Professor, Department of Metallurgical Engineering, is currently working as Dean (Student affairs) at PSG College of Technology. He has obtained his PhD from Anna University, Chennai. His research area includes failure analysis design, Metallurgical property analysis and always do have thrust for newer inventions on material and metallurgical concepts. He has to his credit many national and international Publications. He is an active member IIF, IWS and many other reputed associations in India and abroad.

M. Joseph Malvin Raj is currently pursuing his Post graduation in Engineering Design at Anna University, Coimbatore. His research area includes Alloy development, Processing, Testing, and Characterization of materials and FEA.